\documentclass[journal]{IEEEtran}
\pdfoutput=1
\usepackage{amsmath}
\usepackage{graphicx}
\usepackage{adjustbox}
\usepackage{booktabs}
\usepackage{cite}

\begin{document}

\title{Identification of Hip and Knee Joint Impedance During the Swing Phase of Walking } 

\author{Herman~van der Kooij,~\IEEEmembership{Member,~IEEE},
        Simone S.~Fricke,
        Ronald C.~van 't Veld,
        Ander~Vallinas Prieto,
        Arvid~Q.L.~Keemink,
        Alfred~C.~Schouten,
        and~Edwin~H.F.~van Asseldonk,~\IEEEmembership{Member,~IEEE}

\thanks{
This research was supported by the Dutch NWO Domain Applied and Engineering Sciences (AWARD project, project number: 12850; Reflexioning project, project number: 14903; VICI Flexible robotic suit project, project number 14429.) \emph{(Herman van der Kooij and Simone Fricke are co-first authors.) (Corresponding author: Herman van der Kooij.)}} 
\thanks{H. van der Kooij, S.S. Fricke, R.C. van 't Veld, A. Vallinas Prieto, A.Q.L. Keemink, A.C. Schouten and E.H.F. van Asseldonk are with the Department of Biomechanical Engineering, University of Twente, Enschede, The Netherlands, e-mail: h.vanderkooij@utwente.nl.}
\thanks{H. van der Kooij and A.C. Schouten are also with the Department of Biomechanical Engineering, Delft University of Technology, Delft, The Netherlands}}

\markboth{This is the pre-print of the open access publication in IEEE TNSRE,~Vol. 30,~2022, DOI: 10.1109/TNSRE.2022.3172497}%
{Shell \MakeLowercase{\textit{et al.}}: Bare Demo of IEEEtran.cls for IEEE Journals}

\maketitle

\begin{abstract}
Knowledge on joint impedance during walking in various conditions is relevant for clinical decision-making and the development of robotic gait trainers, leg prostheses, leg orthotics and wearable exoskeletons. Whereas ankle impedance during walking has been experimentally assessed, knee and hip joint impedance during walking have not been identified yet. Here we developed and evaluated a lower limb perturbator to identify hip, knee and ankle joint impedance during treadmill walking. The lower limb perturbator (LOPER) consists of an actuator connected to the thigh via rods. The LOPER allows to apply force perturbations to a free-hanging leg, while standing on the contralateral leg, with a bandwidth of up to 39~Hz. While walking in minimal impedance mode, the interaction forces between  LOPER and the thigh were low ($<$5~N) and the effect on the walking pattern was smaller than the within-subject variability during normal walking. Using a non-linear multibody dynamical model of swing leg dynamics, the hip, knee and ankle joint impedance were estimated at three time points during the swing phase for nine subjects walking at a speed of 0.5 m/s. The identified model was well able to predict the experimental responses for the hip and knee, since the mean variance accounted (VAF) for was 99\% and 96\%,  respectively. The ankle lacked a consistent response and the mean VAF of the model fit was only 77\%, and therefore the estimated ankle impedance was not reliable. The averaged across-subjects stiffness varied between the three time points within 34--66  and 0--3.5 Nm/rad for the hip and knee  joint respectively. The damping varied between 1.9--4.6 and 0.02--0.14 Nms/rad for hip and knee respectively. 
The developed LOPER has a negligible effect on the unperturbed walking pattern and allows to identify hip and knee  impedance during the swing phase. 
\end{abstract}


\IEEEpeerreviewmaketitle

\section{Introduction}

\IEEEPARstart {V}{arious} robotic gait trainers and assistive devices have been developed to overcome neurological disorders affecting walking ability \cite{Esquenazi2017,Morone2017,Meuleman2015}. Humans can walk in various challenging environments and continuously adjust joint impedance. Neurological disorders, e.g. stroke or spinal cord injury, can affect joint impedance and walking ability, due to symptoms like spasticity and hypertonia \cite{Dietz2007,Roy2011,Lee2011a}. Consequently, a detailed understanding of joint impedance during walking, in  people with and without neurological disorders, can improve the design and control of robotic gait trainers and assistive devices \cite{Rouse2014,Maggioni2016}. Further, joint impedance assessment in people with neurological disorders can help improve training protocols and clinical decision making \cite{Maggioni2016}.

Joint impedance is estimated by measuring the response to mechanical perturbations applied to the joint by robotic devices and is often expressed in terms of joint inertia, damping and stiffness \cite{Rouse2014,Lee2015,Huang2020,Koopman2016,Ludvig2017,Zhang1997,Kearney1990}. Joint impedance has been extensively studied and is known to vary with muscle contraction \cite{Hunter1982}, joint position \cite{Weiss1986}, rotation amplitude evoked by the perturbation \cite{Kearney1982} and the velocity (duration) of the applied perturbation. These results imply that joint impedance must vary during movement. Indeed, for the ankle, a time-varying modulation of joint impedance during walking has been reported \cite{Rouse2014,Lee2015}. 

Assessing joint impedance during walking provides additional challenges and requirements for the device applying the mechanical perturbations. An important requirement is that the device should be transparent, i.e. not affect normal walking, when no perturbations are applied. In addition, the device needs to be able to apply the perturbations required for joint impedance estimation. Various devices have been developed with the aim to determine ankle or knee joint impedance during walking \cite{Tucker2017,Yagi2018,Rouse2013a,Roy2009,Andersen2003,Sulzer2009a,Temel2011}. To our knowledge, there are currently no studies that experimentally identified knee or hip joint impedance during walking. For the ankle, a perturbator robot \cite{Rouse2014}  identified the ankle impedance during stance without affecting unperturbed walking. With the wearable Anklebot \cite{Lee2015} the ankle impedance during swing has been identified, but due to its high mass and inertia this device also affects the walking pattern, mainly at the knee and hip.

Here, we 1) developed and evaluated a LOwer limb PERturbator (LOPER) to estimate the hip, knee and ankle joint impedance during walking; 2) developed and validated a new indirect identification method utilizing a personalised rigid-body dynamics model; and 3) obtained a first estimate of  hip and knee joint impedance at different time points during the swing phase in nine able-bodied volunteers.

\section{Device requirements}
The device should be able to apply force perturbations, while effects on the unperturbed walking pattern are negligible. These force perturbations should result in changes in joint angles that can be used to estimate joint impedance with system identification techniques. First, a force bandwidth of 20~Hz is required, based on the torque tracking bandwidth of the LOPES~I that was successfully used to estimate the impedance of the hip and knee during posture tasks \cite{Koopman2016}. Second, the device should not obstruct the range of motion of the leg. Third, we expect that in minimal impedance mode, i.e. when no perturbations are applied, maximal absolute interaction forces of less than 20~N and root mean square (RMS) interaction forces of less than 10~N lead to negligible effects on the walking pattern \cite{Meuleman2013}. Additionally, the root-mean-square difference (RMSD) between joint angles (hip, knee and ankle) with and without the device should be lower than the average within-subject variability while walking.

\section{Methods}
\subsection{Experimental device}
\subsubsection{Design}
\begin{figure}[t]
    \centering
    \includegraphics[trim={0 0.5cm 0 0.7cm},clip]{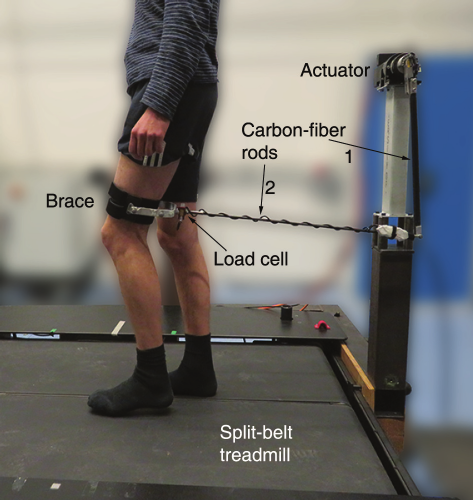}
    \caption{Overview of LOPER (LOwer limb PERturbator) rigidly attached to the frame of a split-belt treadmill. The actuator is connected to the left upper leg of a user with two carbon-fiber rods and a brace with an aluminium frame in between. The rotational motion of the actuator is converted into a linear motion of rod 2. The load cell is placed close to the user (between carbon rod and aluminium frame) and measures the interaction forces between the user and the LOPER.}
    \label{figure:OverviewDevice}
\end{figure}

LOPER consists of a motor, two carbon fiber rods, an aluminium frame and a brace, which is connected to the left upper leg of a human walking on a treadmill  (Fig. \ref{figure:OverviewDevice}). The motor (SMH60, Parker, USA) is attached to the frame of a split-belt treadmill (custom Y-Mill, Forcelink, The Netherlands) with a steel support structure. The first carbon fiber rod (1 in Fig.~\ref{figure:OverviewDevice}, 0.45~m long) is rigidly attached to the motor shaft and is connected to the second rod (2 in Fig. \ref{figure:OverviewDevice}, 0.84~m) via a ball joint linkage. A brace to attach the device to the leg is connected to the rods with an aluminium frame in between. The brace is fixed with Velcro straps on the upper leg, just above the knee. A load cell (FUTEK FSH00086, USA) is implemented between the second rod and the aluminium frame to measure interaction forces.

We have chosen a design where the actuator is not placed on the human to minimize the additional load on the user. Moreover, the choice to place the load cell close to the human limb can result in lower interaction forces and a lower influence of the device on the walking pattern compared with placing the sensor farther away from the user \cite{Zanotto2013a}. Finally, the design allows for sufficient freedom of movement in both sagittal and frontal planes. The chosen rod lengths do not limit the range of motion in the sagittal plane for a person walking on the treadmill. The connection between the aluminium frame and the brace allows for free rotation of the leg in the sagittal plane. The ball joint linkage between the two rods allows for freedom of movement in the frontal plane.

\subsubsection{Hardware and software setup}
The system to control the LOPER and record the force data is composed of one master PC and six slave devices: 
\begin{enumerate}
    \item[1:] a servo drive unit (MOOG MSD 3200 Servo Drive, USA) that controls the velocity of the motor
    \item[2:] an amplifier (MOOG PC CB79047-401A\_HCU, USA) for the signals from the load cell
    \item[3--6:] four Beckhoff modules (1x Beckhoff EK1100, 2x Beckhoff EL3008 and 1x Beckhoff EL4134, Beckhoff Automation GmbH, Germany) which are used to acquire 3 degrees of freedom (DoFs) ground reaction forces and moments from the treadmill and to send a synchronization signal to the motion capture system.
\end{enumerate}
This network of one master and six slave devices is controlled using the EtherCAT real-time control protocol, which runs through EtherLab. On the master PC, a compiled Simulink model (Matlab 2016b, Mathworks, US) runs at 1000~Hz by EtherLab, using TestManager and DLS (Beckhoff Automation GmbH, Germany), to control the device, acquire and save the data from the slaves.

\subsubsection{Control}
\begin{figure*}[t]
    \includegraphics[trim={0 0cm 0 0 cm},clip]{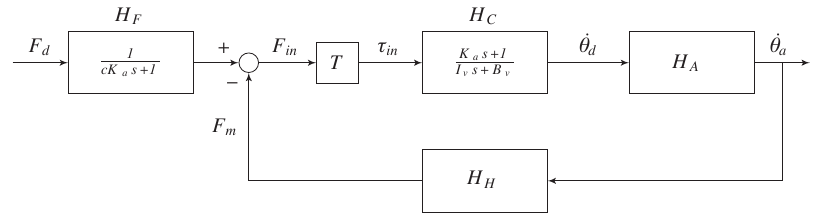}
    \centering
    \caption{Overview of the control scheme. The admittance model ($H_C$) computes the desired angular velocity ($\dot{\theta}_d$) based on the torque input ($\tau_\text{in}$). This torque is obtained by multiplying the force input ($F_\text{in}$) with the moment arm in $T$. $H_A$ represents the actuator and rod dynamics and $H_H$ are the post sensor dynamics, e.g. human and brace. $F_m$ represents the interaction force measured by the load cell and $F_d$ is the desired perturbation force, which is first low-pass filtered through $H_F$.}
    \label{figure:Controller}
\end{figure*}

The device is controlled by a modified admittance controller ($H_C$, Fig. \ref{figure:Controller}) designed to minimize the interaction forces ($F_{in}$) in absence of applied perturbations and track the desired interaction forces ($F_{d}$) of the applied perturbations. The admittance model generates the input ($\dot{\theta}_d$) to the black-box velocity controller, implemented in the servo drive unit.

To minimize interaction forces, an admittance control law with low virtual impedance is implemented. However, a regular admittance controller with low inertia and damping yielded unstable behavior, while interacting with humans. Therefore, the $K_as+1$ term is included in the numerator of $H_C$, which behaves similarly to the acceleration feed-forward described by Keemink et al. \cite{Keemink2018}, and assures interaction stability and faster force convergence. We included this term in the admittance model, because the black-box low level controller of the motor does not accept feed-forward inputs. The perturbations are low-pass filtered ($H_F$) to prevent overshoot, at the expense of frequency bandwidth of the perturbation. When tuning the parameters, first low values for $I_v$ and $B_v$ were selected resulting in fast dynamics. Then, $K_a$ and $c$ were tuned to achieve a stable system with sufficient bandwidth. Tuning resulted in the following values for the parameters: $c=0.5$, $K_a=0.017$~s, $I_v=0.2$~kgm\textsuperscript{2}, $B_v=3$~Nms/rad.

\subsubsection{Safety}
Five features are implemented to assure safety of the user. First, a PVA (position, velocity, acceleration) limiter prevents that the chosen safety bounds for velocity ($\pm$4.71 rad/s) and acceleration ($\pm$500  rad/s\textsuperscript{2}) are exceeded \cite{Meuleman2015}. Close to the position safety bound ($\pm$ 1.22 rad), the PVA limiter limits the actuator velocity resulting in a constant deceleration. Beyond this position bound, the velocity is limited to 0 m/s in the direction away from the neutral position. Second, the outputs from the sensors (encoders and load cells) are continuously checked and the motor is switched off directly when a threshold for either position ($\pm$1.31 rad), velocity ($\pm$6.28 m/s) or actuator torque (120 Nm) is reached. These boundaries are slightly larger than the PVA limiter bounds to prevent that the motor turns off each time a slightly larger value is reached. Third, when the servo drive unit detects an angle larger than $\pm$1.34 rad the motor is switched off directly. Fourth, two emergency buttons are placed close to both experimenter and user, which directly switch off the motor when pressed. Fifth, users wear a safety harness to prevent falls.

\subsection{Experimental evaluation}
\subsubsection{Ethics}
The experiments were approved by the ethics committee of the Electrical Engineering, Mathematics and Computer Science faculty of the University of Twente (approval number: RP 2019-83). All participants gave informed consent prior to the experiments.

\subsubsection{Device performance in static situations}
The performance of the LOPER was evaluated in two tests in which a participant (1 male, 24 years, height 1.84 m, weight 66 kg) was standing on his fully extended right leg, while the left leg, connected to the LOPER, was lifted from the ground and relaxed, i.e. free-hanging. In the first test, the force bandwidth was evaluated using a force perturbation of filtered white noise for 60 s (cut-off frequency 60 Hz, peak-to-peak amplitude of 60~N). In the second test, step responses were evaluated for force perturbations with several amplitudes (20, 40 and 60 N in forward direction, each applied 10 times).

\subsubsection{Device performance during walking}
Five participants (4 male, 1 female, 26.4$\pm$1.3 years, height 1.71$\pm$0.09 m, weight 68.4$\pm$11.5 kg) without any self-reported impairments in their lower limbs walked two times four minutes on the treadmill to assess the performance of the minimal impedance mode of the device. Participants walked on the treadmill at 0.5 m/s while following a metronome (36 strides/min) to keep the stride frequency as constant as possible. The low walking speed was chosen for all experiments as it is relevant for clinical applications in people with neurological disorders. The stride frequency was chosen based on the average stride frequency during pilot experiments in two able-bodied participants walking at 0.5m/s. In the first trial, participants walked without the device. In the second trial, participants walked with the LOPER in minimal impedance mode.

\subsubsection{Assessing joint impedance during swing phase}
Nine participants (6 male, 3 female, 26.1$\pm$1.2 years, height 1.73$\pm$0.14 m, weight 64.9$\pm$12.1 kg) without any self-reported impairments in their lower limbs were included in the joint impedance assessment experiments. To estimate joint impedance during the swing phase, force perturbations were applied at three onset times: 50, 175 and 300 ms after toe-off. These onset times were chosen to cover the entire swing phase. Rectangular pulses with a pulse width of 100 ms and an amplitude of 40 N (forward) were applied as perturbation in all cases. In order to time the perturbations appropriately, a phase detection algorithm was used to detect toe-off based on the vertical ground reaction forces. The perturbation experiments were split into six trials, two trials for each perturbation onset time, with the trial order randomized for each participant. Each trial lasted four minutes, with a walking speed of 0.5 m/s. Participants followed a metronome (36 strides/min). For each trial, the perturbations were applied randomly after every 3--5 strides during the swing phase of the left leg at the specified perturbation onset time.

\subsection{Data recording and processing}
During all experiments, the load cell between the rod and leg brace measured the interaction forces. For the experiments during walking, 3 DoF ground reaction forces and moments were recorded from the split-belt treadmill (Motek Medical, Houten, the Netherlands). A motion capture system (Qualysis AB, Sweden) recorded the movements of the participants and the brace of LOPER. Eight Oqus 600+ cameras were placed around the treadmill and marker movements were recorded at 128~Hz. The synchronisation signal, that was sent from the LOPER setup to the motion capture system, was recorded at 1024~Hz using an analog to digital conversion board. Two markers were placed on the brace of the device to be able to calculate the angle of the brace and the direction of the applied forces. A total of 32 markers were placed on bony landmarks and body segments of the participants to record kinematics of feet, lower legs, upper legs, pelvis and trunk. All data was processed in Matlab 2018b. In the following sections and figures, a positive sign indicates a force (of the device) in forward direction, i.e. walking direction, a hip and knee flexion angle/torque, and an ankle dorsiflexion angle/torque. 

\subsubsection{Device performance in static situations}
The frequency response function (FRF) of the force controller (see Fig. \ref{figure:Controller}) was determined   by dividing the following cross-spectral densities ($S_{xy}$):
$$H_{{F_d}\text{to}{F_m}}=\frac{S_{F_m F_d}}{S_{F_d F_d}}$$
where $F_d$ is the input perturbation and $F_m$ the measured interaction force. Welch averaging with a Hann window (size: 5000 samples, overlap: 50 samples) was used to calculate these spectral densities. The bandwidth of the force controller was calculated as the -3~dB point of the FRF.

Step responses were averaged across the ten repetitions to reduce noise. Rise time was calculated for each perturbation amplitude and defined as the time needed by the measured interaction force to rise from 10\% to 90\% of the steady-state response. Percentage overshoot was determined relative to the steady state response.

\subsubsection{Device performance during walking} 
Gait phase estimates, calculated based on ground reaction forces and moments, were used to cut the LOPER data into strides, i.e. a full gait cycle. For the walking trials without perturbations, interaction forces between the device and the human were averaged over all strides within a participant. The RMS interaction forces and maximal absolute interaction forces were determined for each participant and averaged across participants.

Joint and segment angles were determined based on the motion capture data. The measured marker positions were filtered in Matlab 2018b with a 4\textsuperscript{th} order zero-phase 40~Hz low-pass Butterworth filter. In OpenSim 4.0 the gait2392 model was used to perform inverse kinematics. The resulting joint angles, which are called `measured joint angles' in the remainder of this article, were cut into strides similar to the LOPER data. For each data point in a stride, we considered it to be an outlier if the value of this data point was 1.5 times the interquartile range below the first quartile or above the third quartile. The interquartile ranges were determined from all measured strides at the same relative gait cycle times. If more that 20\% of the data points in a stride were classified as an outlier, the complete stride was discarded. After outlier removal, strides were averaged within each participant. For each participant, the RMSD between the average joint trajectories for the trials with and without the device were calculated. TheRMSD was compared to the average intra-subject variability (ISV\textsubscript{av}) defined as twice the average standard deviation ($\sigma_{i_p}$) between strides within a participant \cite{Meuleman2013}: $$\text{ISV}_\text{ave}=\frac{2}{n_p}\sum_{i_p=1}^{n_p}\sigma_{i_p.}$$ where $i_p$ indicates the subject number and $n_p$ the total number of subjects. We assumed that effects of the device were negligible if the RMSD between the trial without and with the device was smaller than the ISV\textsubscript{ave} for the trial without the device \cite{Meuleman2013}.

\subsubsection{Assessing joint impedance during the swing phase} 
\begin{figure}[t]
    \includegraphics{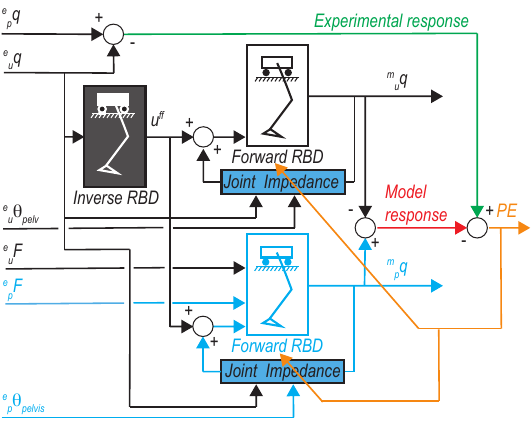}
    \caption{Schematic overview of indirect identification method to identify joint impedance, which uses a rigid-body dynamics (RBD) model of the swing leg. Joint impedance is found by minimisation the prediction error (PE) that is the difference in model response (red line) and experimental response (green line). The symbol $q$ refer to the kinematics, $u_\text{ff}$ to the generalised feed-forward forces, $\theta_\text{pelvis}$ to the pelvis angle, and $F$ the forces exerted by the LOPER device. The superscript $e$ and $m$ refer to experimental and model states, respectively. The subscripts $u$ and $p$ refer to unperturbed and perturbed states, respectively. The model and data used for the unperturbed and perturbed are color-coded with black and blue lines, respectively.}
    \label{figure:IndentificationMethod}
\end{figure}

We developed an indirect identification method to identify the joint impedance, which makes use of inverse and forward dynamical models of the swing leg dynamics (Fig. \ref{figure:IndentificationMethod}). The underlying assumption made is that limb motion was driven by feed-forward (ff) and feedback (fb)  control and that the unperturbed kinematics is the result of feed-forward control only. The feedback pathway consists of the joint damping and stiffness (i.e. the joint impedance) that forces the swing leg back to its nominal or desired pathway, i.e. the measured motion of the unperturbed trials. Using the experimental kinematics from the unperturbed trials, the feed-forward generalised forces ($u^\text{ff}$) are obtained from inverse dynamics. They are used as input for the forward simulation of the model for the unperturbed and perturbed conditions. From the forward model simulations a model response to forces exerted by the LOPER device can be obtained. The difference between the model and experimental response determines the prediction error (PE), which is used to identify the unknown joint stiffness and damping by numerical minimisation of the PE. 

Experimental joint angles were determined and processed as described in the previous paragraph. Interaction forces were filtered with the same 4\textsuperscript{th} order zero-phase 40~Hz low-pass Butterworth filter, after which the interaction forces were resampled to 128 Hz. 

Joint angles and applied LOPER forces were cut into strides. For each perturbed step, the most similar unperturbed stride was found, based on the RMSD between the perturbed and unperturbed strides for the last 25 ms before the perturbation onset time, and subtracted from the perturbed trial. For the impedance estimation (see below), an analysis window was used that included 250 ms after the perturbation onset time. For the last perturbation onset time (300 ms), strides with a swing time shorter than 550 ms (300 ms + 250 ms analysis window) were removed to avoid the inclusion of the stance phase in the analysis. Again, outliers were removed as described above, but only taking into account the range of data points that was used for joint impedance estimation (25 ms before to 250 ms after the perturbation).

After removing outliers, average joint angles and applied LOPER forces were used to estimate the joint impedance of the hip, knee and ankle. The analysis of each perturbation onset time included data from 25 ms before to 250 ms after the perturbation. 

\begin{figure}[t]
    \includegraphics{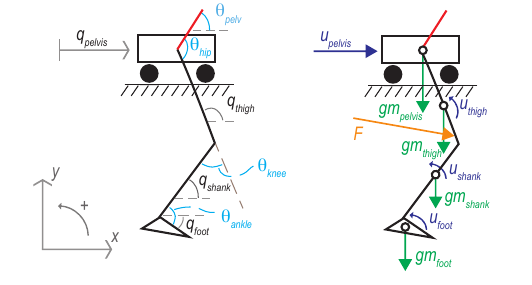}
    \caption{Graphical presentation of the rigid-body dynamics model used to predict the response of the perturbation on swing leg kinematics. Left: Definition of the segment and joint angles. The generalised coordinates ($q$) and the given segment angle of the pelvis ($\theta_{pelv}$) are defined in the global reference frame. The hip ($\theta_{hip}$), knee ($\theta_{knee}$), and ankle ($\theta_{ankle}$) angles are defined in local coordinate systems as relative angles between segments. The pelvis orientation (red segment) is part of the rigid-body dynamics but used as independent input parameter to be able to calculate the hip angle. Right: The modelled forces and torques acting on the system, which are the gravitional forces ($gm$), the generalized forces ($u$), and the force ($F$) excerted by the LOPER device.}
    \label{figure:RGB model}
\end{figure}

To identify the joint impedance of the leg in swing we used a 2D-model of the rigid body dynamics (RBD) consisting of a triple pendulum hanging on a cart that could only move forward and backward. The triple pendulum represents the foot, shank and thigh, whereas the cart represents the horizontal pelvis motion (see Fig. \ref{figure:RGB model}).

The equations of motion of this model are:
$$
M(q)\ddot{q}=-C(q,\dot{q})+G(q)+u+J^T(q)F
$$
where $M$ is the mass matrix, $C$ the vector with centrifugal and Coriolis forces, $G$ the vector with gravitational forces, $F$ the external force vector from the LOPER device, $J$ the Jacobian that relates the point of interaction to the generalised coordinates $q$, and $u$ the vector with generalised forces. The latter two are defined (see also Fig. \ref{figure:RGB model}.) as:
$$q=\left\{ q_\text{pelvis},q_\text{thigh},q_\text{shank},q_\text{foot} \right\}^T$$
$$u=\left\{ u_\text{pelvis},u_\text{thigh},u_\text{shank},u_\text{foot} \right\}^T$$
For the foot, shank and thigh, mass, inertia, length and location of the centre of mass were exported from OpenSim for each individual. For the cart the only relevant parameter is its mass that was defined as the the difference of total body mass and mass of one swing leg. The mass of the cart thus equals the reflected mass of the rest of the body to the swing leg. The cart model captures the interaction of the rest of the body with the swing leg dynamics. Ignoring this would result in incorrect estimates of the generalised forces using inverse dynamics (see later). We did not consider the vertical motions of the pelvis (cart) since the vertical accelerations of the pelvis were much smaller than its horizontal accelerations. Since only accelerations are relevant, as they excite the motion of the swing leg (and not the velocity or position), this simplification can be justified.

 The feedback torques from the joint impedance are defined as the result of the joint stiffness ($K$) and damping ($D$) multiplied with the differences between joint angles and joint angular velocities from the model and the experimental kinematics from the unperturbed trials:
$$
\begin{aligned}
T_\text{hip}^{\text{fb}}={} &  -K_\text{hip}\left( ^{m}\theta_\text{hip}-{^{e}_u\theta_\text{hip}} \right) \\ & -D_\text{hip}\left(^{m}\dot{\theta}_\text{hip}-{^{e}_u\dot{\theta}_\text{hip}} \right)
\end{aligned}
$$

$$
\begin{aligned}
T_\text{knee}^{\text{fb}}={} &  -K_\text{knee}\left( ^{m}\theta_\text{knee}-{^{e}_u\theta_\text{knee}} \right)  \\& -D_\text{knee}\left(^{m}\dot{\theta}_\text{knee}-{^{e}_u\dot{\theta}_\text{knee}} \right)
\end{aligned}
$$

$$
\begin{aligned}
T_\text{ankle}^{\text{fb}}={} &  -K_\text{ankle}\left( ^{m}\theta_\text{ankle}-{^{e}_u\theta_\text{ankle}} \right)  \\& -D_\text{ankle}\left(^{m}\dot{\theta}_\text{ankle}-{^{e}_u\dot{\theta}_\text{ankle}} \right)
\end{aligned}
$$
where the subscripts $u$ refer to the kinematics of the unperturbed trials, the superscripts $e$ express the kinematics are from experimental data, and the superscripts $m$ denote that these are model variables. 

The joint angles are defined as function of the generalised coordinates and the pelvis angle that is not a model variable but an independent input (from measurements), given by:
$$\theta_\text{hip}=\theta_\text{pelvis}-q_\text{thigh}$$
$$\theta_\text{knee}=q_\text{thigh}-q_\text{shank}$$
$$\theta_\text{ankle}=q_\text{shank}-q_\text{foot}$$
Consequently the joint torques can be converted into the generalised forces by:
$$u_\text{thigh}=T_\text{knee}-T_\text{hip}$$
$$u_\text{shank}=T_\text{ankle}-T_\text{knee}$$
$$u_\text{foot}=-T_\text{ankle}$$

We assumed that the motion of the unperturbed trials did originate completely from feedforward control. The corresponding feed-forward generalised forces can be obtained from the inverse rigid body dynamics using the experimental data of the unperturbed trials:
$$
u^{\text{ff}}=M\left( _u^{e}q \right){_u^{e}\ddot{q}}+C\left(_u^{e}q,{^{e}_u\dot{q}} \right)
-G\left( _u^{e}q \right)
$$
Note that $u^{\text{ff}}$ also includes the forces from the LOPER device during unperturbed walking, mapped to the generalised forces.

To predict the response we simulated the response of the unperturbed and perturbed conditions by numerical integration (ODE45, Simulink, Matlab 2020a, Mathworks, US)  of the second order differential equations of the forward body dynamics:
$$
\begin{aligned}
_u^{m}\ddot{q}= M^{-1}\left(_u^{m}q  \right) {} & \{ -C(_u^{m}q,{_u^{m}\dot{q}})+G(_u^{m}q) \\ & +u^\text{ff} +u^\text{fb}(_u^{m}q,{_u^{e}q}) \}
\end{aligned}
$$

$$
\begin{aligned}
_p^{m}\ddot{q}= M^{-1}\left(_p^{m}q  \right) {} & \{ -C(_p^{m}q,{_p^{m}\dot{q}})+G(_p^{m}q) \\ & +u^\text{ff}+u^\text{fb}(_p^{m}q,{_u^{e}p}) \\&  +J^{T}(_p^{e}q){_p^{e}F}-J^{T}(_u^{e}q){_u^{e}F} \}
\end{aligned}
$$
where the subscript $p$ denotes the perturbed kinematics. The simulation of the unperturbed dynamics is driven by feed-forward and feedback torques from the joint impedance. Ideally the feedback torques will be zero, but due to small  inconsistencies between the numerical differentiation in the inverse model and numerical integration in the forward model small feedback torques occur. The simulation of the perturbed dynamics is driven by feed-forward and feedback torques from the joint impedance, plus the LOPER forces mapped to the generalised forces from the unperturbed and perturbed trials. 
From these forward model simulations we can calculate the predicted response and compute the difference with the experimental response, which is the prediction error: 

$$
\text{PE}=\left(  {_p^{e}q}-{_u^{e}q} \right)-\left(  {_p^{m}q}-{_u^{m}q} \right)
$$

The unknown parameters of the joint stiffness and damping were found by numerical optimisation. We used  a non-linear least squares optimisation (Matlab 2020a, Mathworks, US, lsqnonlin) that minimises sum of the square of PE. This prediction error quantifies how well the feedback model can describe the experimentally observed differences between perturbed and unperturbed trials. Minimisation of this error will result in those feedback parameters that best describe the human response to the perturbations. Instead of the generalised coordinates we used the hip, knee and ankle angles put in one vector to calculate the PE. To avoid the chance to end up in a local minimum, each optimisation was repeated ten times and the solution with the best fit was taken. Each optimisation was started from an initial guess randomly chosen between the parameters bounds. The lower parameter bounds were set to zero. The upper bounds were set to 200~Nm/rad and 10~Nms/rad for $K$ and $D$, respectively.
How well the model fitted the experimental data was expressed by the variance accounted for (VAF) where 100\% represents a perfect fit.

To validate our new joint stiffness and damping identification method, we used synthetic data instead of experimental data. To this end the forward RBD model was simulated with different values of joint stiffness and damping. The optimisation procedure as outlined above was employed to identify the known $K$ and $B$. If the correct $K$ and $B$ will be found by the optimisation procedure, the corresponding PE will be zero and the VAF 100\%. Still, it might be possible that also other combinations of $K$ and $B$ result in a zero PE, which would result in an incorrect identification of the joint impedance. To generate the synthetic data the model stiffness for each separate joint could be 0, 75 or 150 Nm/rad, and the model joint damping could be 0, 2, 4 Nms/rad. We evaluated all possible combinations of joint stiffness and damping, resulting in 729 combinations. To make the validation more realistic we validated the identification without and with synthetic measurement noise that was added to the synthetic perturbed and unperturbed ‘experimental’ generalised coordinates. The noise had a standard uniform distribution  with peak to peak levels of 0.01 rad or m, and all noises added were uncorrelated. To evaluate how well the identification method performed we reported the maximum, minimum and standard deviation of the parameter estimation errors.

\section{Results}
\subsection{Device performance in static situations}
\begin{figure}[t]
    \includegraphics{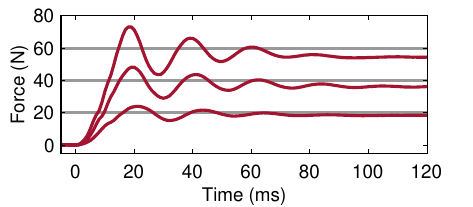}
    \caption{Step responses (averaged over 10 responses) in one participant with three amplitudes (20, 40, 60 N) that were applied to a free-hanging leg in forward direction. The participant was standing on his fully extended contralateral leg. The grey lines represent the desired steady state values.}
    \label{figure:StepResponse}
\end{figure}

We designed and evaluated the LOPER, which can be used to apply force perturbations to the left upper leg. A bandwidth of 39~Hz was found for the FRF (input $F_d$, output $F_m$, Bode plot not shown) when applying perturbations to a free-hanging leg, which is higher than the required bandwidth of 20~Hz. Step responses show rise times of 9.7--10.3 ms for 20, 40 and 60 N perturbations (Fig. \ref{figure:StepResponse}). An overshoot of 29--32\% of the steady state responses was found. The steady state responses did not fully reach the desired values, e.g. the steady state response for the 40~N perturbation was 36~N (Fig. \ref{figure:StepResponse}).

\subsection{Device performance during walking}
\begin{figure}[t]
    \includegraphics{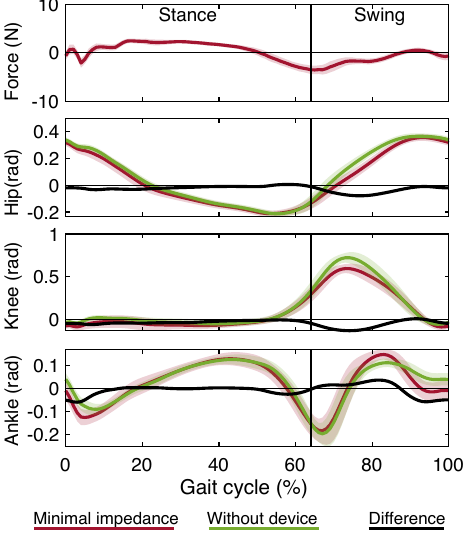}
    \caption{Interaction forces and joint angles for the left leg while walking with the LOPER in minimal impedance mode and without the device for a representative participant. The positive axis shows a force in forward direction and a (dorsi) flexion angle, respectively. The shaded areas illustrate the standard deviation across strides. Left heel strike occurred at 0\% of the gait cycle.}
    \label{figure:AnglesInteractionForces}
\end{figure}

To evaluate the performance of the device during unperturbed walking, participants walked without the LOPER, and with the LOPER attached to their left upper leg in minimal impedance mode. Due to outlier removal about 1.7\% and 1.1\% of the data was excluded from analysis for the with LOPER and without LOPER condition, respectively. In line with the requirements, low interaction forces were found in minimal impedance mode (Fig. \ref{figure:AnglesInteractionForces}, top). The average ($\pm$~standard deviation across participants) RMS interaction forces in stance and swing phase were 1.97$\pm$0.21 N and 2.13 $\pm$0.30 N and the maximal absolute interaction forces were 4.46$\pm$0.82 N and 4.63$\pm$0.79 N. The largest interaction forces were found at the end of the stance phase/beginning of the swing phase, which can be attributed to the high acceleration of the leg at this phase of the gait cycle.
On average the differences in the hip, knee and ankle angles between walking with the device in minimal impedance mode and without the device (Fig. \ref{figure:AnglesInteractionForces}) are smaller than the variability within subjects.
During stance, for the hip, knee and ankle, the average RMSD between the trial without the device and the minimal impedance trial (hip: 0.024$\pm$0.014~rad, knee: 0.032$\pm$0.017~rad, ankle: 0.017$\pm$0.008~rad) was lower than the ISV\textsubscript{ave} (hip: 0.040~rad, knee: 0.062~rad, ankle: 0.040~rad) for walking without the device. Also during swing, for the hip, knee and ankle, the average RMSD between the trial without the device and the minimal impedance trial (hip: 0.035$\pm$0.014~rad, knee: 0.071$\pm$0.045~rad, ankle: 0.035$\pm$0.011~rad) was lower than the ISV\textsubscript{ave} (hip: 0.059~rad, knee: 0.129~rad, ankle: 0.064~rad) for walking without the device. The largest differences between the trial with and without the device were found during initial and mid-swing for all joints, probably due to the larger interaction forces at end of the stance and the beginning of the swing phase (Fig.~\ref{figure:AnglesInteractionForces}, top panel).

\subsection{Validation of joint impedance identification method}
For each combination (in total 729 combinations) of the possible values of joint stiffness and damping, the majority of the ten optimisations for each combination converged to the same global minimum of the squared 2-norm of the PE (results not shown). The solution(s) with the lowest 2-norm of the PE  were taken.
In case no synthetic measurement noise was added, the parameter estimation errors were between -0.87 and 0.59 Nm/rad and between  -9.2$\cdot$10\textsuperscript{-2} and 4.7$\cdot$10\textsuperscript{-2} Nms/rad, for the joint stiffness and damping respectively. 
When adding the synthetic measurement noise, the estimation errors increased (see Table~\ref{table: Estimation Errors}). The knee estimates were the least sensitive to adding noise. For the hip and knee, errors in stiffness estimates were less than 4\% and in damping estimates less than 14\% of the range of explored stiffness and damping values, respectively. But the minimal and maximal estimation error for the ankle stiffness and damping were considerably larger. 

\begin{table}[t]
\centering
\caption{Parameter estimation errors from synthetic data from model simulation with additive measurement noise}
	\begin{adjustbox}{width=\columnwidth}
		\begin{tabular}{@{} l  r  r  r r r  r @{}}
			\toprule
			-	  		& $K_\text{hip}$		& $D_\text{hip}$ 				& $K_\text{knee}$ 		& $D_\text{knee}$ 	& $K_\text{ankle}$ 		& $D_\text{ankle}$	\\ 
			& (Nm/rad) & (Nms/rad)	& (Nm/rad)	& (Nms/rad)	& (Nm/rad)	& (Nms/rad)			\\ 
						
								\\ \midrule
			Min. Error	& -6.2   & -0.57   	& -2.5   	 
						& -0.11   & -120   	& -4   				 \\ \hline
			Max. Error 	& 6.5  	& 0.50 		& 3.5   
						& 0.19   & 120   	& 10   				 \\ \hline
			Std. Error 	& 15  	& 0.12   	& 0.53  	
						& 0.030   & 24   	& 1.5   				 \\ \hline
		\end{tabular}
	\end{adjustbox}
	\label{table: Estimation Errors}
	\vspace{-0.5cm}
\end{table}

\subsection{Assessing joint impedance during swing phase}
\begin{figure}[t]
    \includegraphics{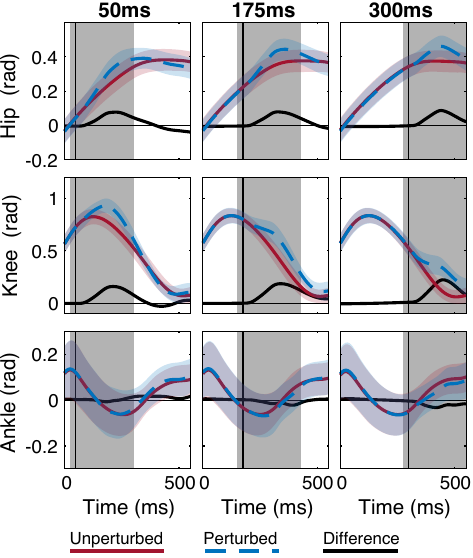}
    \caption{Measured joint angles for the perturbed and unperturbed steps and the difference between them during the swing phase (0~ms is at the start of the swing phase). Each column shows the results for one perturbation onset point and the window used to fit the model is indicated with a grey box. The vertical line indicates the beginning of the perturbation.}
    \label{figure:SwingTrajectories}
\end{figure}

To assess joint impedance during walking, participants walked six times on the treadmill, while perturbations were applied with the LOPER at initial (50--150~ms), mid- (175--275~ms), and terminal swing (300--400~ms) (Fig. \ref{figure:SwingTrajectories}). All participants performed two trials for each perturbation onset time. However, for one participant, one trial (175~ms) is missing due to data acquisition issues, leaving one trial for analysis. Due to outlier removal about 4.5\% and 4.2\% of the data was excluded from analysis for the unperturbed and perturbed steps, respectively. For the last perturbation onset time (300~ms), 38\% of the total amount of steps for all participants had to be removed, because swing times were shorter than 550~ms. Still, at least 30 steps were included for the analysis for each participant at each perturbation onset time.

\begin{figure}[t]
    \includegraphics{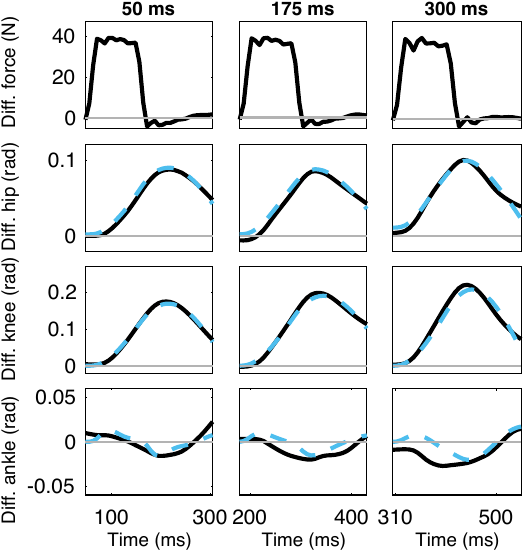}
    \caption{Comparison between experimental data (solid lines) and model predictions (dashed lines) of the differences between perturbed and unperturbed steps in forces exerted by the LOPER device and the hip, knee,  and ankle joint angles from a typical subject. On the x-axis, 0ms is at the start of the swing phase and only the analyzed window (grey box in Fig. \ref{figure:SwingTrajectories}) is shown for each perturbation onset point.}
    \label{figure:FitTrajectories}
\end{figure}

The applied perturbations led to more hip and knee flexion during and after the perturbation irrespective of when it was applied (Fig. \ref{figure:SwingTrajectories} and \ref{figure:FitTrajectories}). At the ankle the response was smaller and not consistent. As soon as the perturbation was applied, differences between perturbed and unperturbed increased until around 150~ms after which the differences decreased again and returned to zero around 400~ms after the perturbation.

\begin{figure}[t]
    \includegraphics{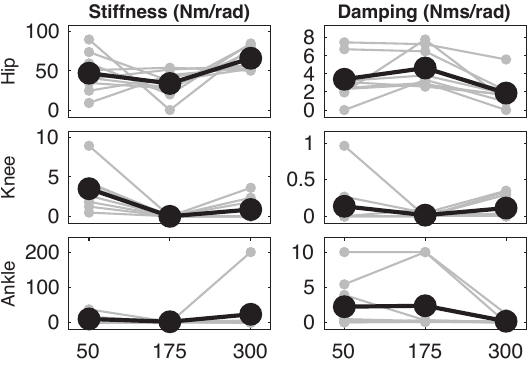}
    \caption{Estimated Stiffness ($K$) and damping ($B$) of the hip, knee and ankle joints. The black line with markers indicates the average across all participants. Each grey line with markers shows the results for one participant.}
    \label{figure:EstimatedParameters}
\end{figure}

For all individuals and all conditions the identified model could predict the response in the hip and knee joints well, whereas for the ankle the model fit was less good  (see Fig.~\ref{figure:FitTrajectories}). The goodness of the model fits are also reflected in the VAF, which ranged between 96.5--99.9\%, 87.3--99.5\%, and 18.6--97.7\% for the hip, knee and ankle joint respectively for the different subjects and time points. The averaged VAF across subjects and three time points, were 99.0\%, 95.8\%, and 77.8\% for these joints. Overall these VAF metrics show that the model is able to capture the responses at the hip and knee very well. For the ankle, the experimental response is much smaller and inconsistent. Consequently the model fit is worse.

For the estimated joint stiffness and damping, some variability was found between participants and perturbation onset times (Fig.~\ref{figure:EstimatedParameters}). On average the lowest hip and knee joint stiffness was found for a perturbation onset at 175~ms. The estimated hip stiffness and damping were larger than in knee and ankle. The estimated stiffness and damping in the knee was the lowest.

\section{Discussion}
The goal of our study was three-fold: 1) to develop and evaluate a lower limb perturbator (LOPER) to estimate the apparent swing leg joint stiffness during walking; 2) to develop and validate a new identification method accounting for the dynamical interactions between different body segments; and 3) to obtain a first estimates of hip joint and knee impedance at different time points during the swing phase of walking in able-bodied volunteers.

\subsection{Performance of the device}
The LOPER device we developed fulfills all requirements needed to identify joint impedance during the swing phase. LOPER obtained a force bandwidth of 39 Hz when forces were applied to a free-hanging leg while the participant stood on the contralateral leg, which is higher than the required 20 Hz. The mechanical design, where forces are transmitted to the upper leg using rods, did not restrict the motions of the hip, knee and ankle during walking. Moreover, thanks to the high torque tracking bandwidth and subsequently low interaction forces in unperturbed walking, the effect of the LOPER in minimal impedance mode on gait kinematics was smaller than the within subject variability. Walking with the LOPER resulted in a more gradual swing, i.e. less acceleration in the beginning of the swing phase. In the future, controllers that take into account the cyclic behaviour of walking could be used to further increase transparency \cite{VanDijk2013} and reduce this effect.

\subsection{Validity of methods and study limitations}
The novelty of our method is that we apply one force to the limb that excites the hip, knee and ankle joint. Existing methods apply joint torque or a joint position perturbation. Our method exploits the mechanical coupling between the leg segments; the force applied at the thigh excites the system and induces accelerations at all joints, which can be understood by recognizing that the non-diagonal entries of the mass matrix are not zero. The validity of the developed indirect identification method was demonstrated by computer simulations. In the experiments, the applied force evoked a response of approximately 0.1 rad at the hip and 0.2 rad at the knee was elicited, while at the ankle a much smaller and less consistent response was observed. For hip and knee, high VAF values for the fitted model were found but low VAF values for the fitted ankle response. The low VAF values at the ankle are caused by the small response elicited at the ankle, which was also the reason why in the validation by model simulations of the identification method the ankle stiffness and damping could not be estimated accurately when artificial noise was added to the simulated responses, in contrast to the hip and knee.

The underlying assumption of our method is that the unperturbed motion is driven purely by feed-forward control and stiffness and damping are defined as deviations from the unperturbed motion. In case the unperturbed motion is also caused by joint impedance the reference pattern for joint impedance will be different from the unperturbed motion. Also other methods (implicitly) assume that the unperturbed motion is the reference for joint impedance. In our method it is possible  to include alternative joint impedance reference trajectories like those  predicted by the optimal feedback framework \cite{Todorov2004}, also as a method to further experimentally validate this framework.

Our study has several limitation. We applied the force perturbation only in one direction. Since the estimated joint impedance is dependent also on the perturbation direction \cite{Huang2020} it would be recommended to also study the sensitivity of force direction with our method.

Another limitation of our study is that the identification method ignored inter-joint stiffness and damping, which originate from bi-articular muscles. It has been shown that including the inter-joint impedance in the model results in better model fits \cite{Koopman2016}. In the walking conditions as used in this study it is unlikely that including these effects will much improve the already high VAF values we found. Including more model parameters might also require additional independent perturbations as in \cite{Koopman2016} to enrich the data-set used to uniquely estimate these parameters.

\subsection{Joint impedance during swing phase}
Since this is the first study that determined hip and knee joint impedance during walking we can only compare our results for the hip and knee with results from studies that used other conditions than walking (Table \ref{table: Comparison studies}). Some of these studies also included a position \cite{Koopman2016} or force \cite{Huang2020,Zhang1997} task, which resulted in higher contraction levels and thus joint impedance compared to relaxed conditions. Since in slow walking the leg almost moves ballistically, a comparison with relaxed conditions is the most appropriate. Since our estimate of ankle joint impedance is not reliable, we do note make a comparison for the ankle impedance with \cite{Lee2015}.

\begin{table}[t]
\centering
\caption{Comparison of estimated joint impedance in different studies that used force ($F_{pert}$) or position ($P_{pert}$) perturbations during swing phase of walking (W), static conditions (S) or during imposed motions (I) in a  relax task.}
	\begin{adjustbox}{width=\columnwidth}
		\begin{tabular}{@{} l  r  r  r r r  r @{}}
			\toprule
			Reference	  		& $K_\text{hip}$		& $D_\text{hip}$ 				& $K_\text{knee}$ 		& $D_\text{knee}$ 		\\ 
			& (Nm/rad) & (Nms/rad)	& (Nm/rad)	& (Nms/rad)			
						
								\\ \midrule
			This study ($F_{pert}$,W)	& 34--66   & 1.9--4.6   	& 0--3.5   	 
						& 0.02--0.14    				 \\ \hline
			\cite{Koopman2016} ($F_{pert}$,S+R) 	& 50-60 	& 7-9 		& 15-20   
						& 1-2  			 \\ \hline
			\cite{Huang2020} ($P_{pert}$,S+R) 	& 75--318  	& 2-21 	& --    	
						& --         				 \\ \hline
			\cite{Zhang1997} ($P_{pert}$,S+R) 	& --  	& -- 	& 36-75    	
						& 1.8-2.2        				 \\ \hline			
	 	\cite{Ludvig2017} ($P_{pert}$,I+R) 	& --   	& --    	&  25--30    	
						& --   	 \\ \hline
		\end{tabular}
	\end{adjustbox}
	\label{table: Comparison studies}
	\vspace{-0.5cm}
\end{table}
The estimated joint stiffness and damping we found are overall lower than previously found estimates. These differences can be explained by differences in the identification methods and by differences in experimental conditions. First, we and others \cite{Koopman2016} used force perturbations, while other studies \cite{Ludvig2017,Huang2020} applied fast and small position perturbations. 
Since it is known that estimated stiffness decreases with evoked joint rotation amplitude \cite{Kearney1982}  and perturbation duration \cite{DeVlugt2011}, the shorter and faster perturbations and smaller joint rotations in \cite{Ludvig2017,Huang2020} could be a possible cause for the higher stiffness estimates they found.
Second, our lower estimates of hip and knee stiffness compared to those found in static conditions can also be understood from a previous study in which joint stiffness estimates during movement were considerably lower than when estimated in a position and torque-matched static task \cite{Ludvig2017}.
Finally, in some studies \cite{Huang2020,Ludvig2017} the joint stiffness estimates also include the gravitational stiffness of the leg and braces attached to the leg, which results in an overestimation of the biological joint stiffness. In \cite{Zhang1997} the gravitational stiffness estimate for the knee  was 5.9 Nm/rad. For the subjects included in our study, the gravitational stiffness was approximately 35 Nm/rad for the hip.

\subsection{Future directions}
In the future, the device and analysis methods should be extended to address the aforementioned limitations. For this purpose, another push-pull rod just above the ankle can be added to excite the ankle joint more. Multi-input multi-output (MIMO) techniques could then also be used to determine joint impedance as well as inter-joint impedance, e.g. due to biarticular muscles \cite{Koopman2016}. Also applying position perturbations as in \cite{Ludvig2017} instead of force perturbations could be considered since they need a shorter time window to estimate joint stiffness.

Furthermore, our methodology could be extended to the stance phase. This results in additional requirements for the device and analysis methods. First, larger forces need to be applied during the stance phase to result in similar changes in joint angles, e.g. due to weight bearing and an expected larger joint stiffness in the stance phase \cite{Lee2016}. Second, perturbations during stance phase might have larger influences on the walking pattern, e.g. disturbing balance. To estimate the joint impedance during the stance phase, the RBD model needs to be adopted accordingly.

The presented method allows for identifying joint impedance in people with neurological disorders, e.g. stroke, spinal cord injury, to gain a better understanding of their impaired walking ability. However, applying these methods in people with neurological disorders will pose additional challenges. People with neurological impairments can have problems following the metronome leading to a larger gait variability \cite{Roerdink2009}. This larger gait variability can increase the difficulty of timing the perturbations consistently and could influence the responses to perturbations. In addition, impairments, e.g. spasticity or hypertonia, can result in an increased joint stiffness \cite{DeVlugt2010,Roy2011}. Therefore, larger forces might be needed to bring about clear changes in joint angles in people with neurological disorders.

\section{Conclusion}
We developed a lower limb perturbator (LOPER) with a high force tracking bandwidth and with an effect on the unperturbed walking pattern that was smaller than the within subject variability. The developed identification method was able to predict the responses at the hip and knee  with high accuracy as presented by the high VAF values we found. With the LOPER device, we were able to obtain first estimates of hip and knee joint impedance during walking.

\bibliographystyle{IEEEtran}
\bibliography{Papers-JointImpedance.bib}
\end{document}